\documentclass{bmvc2k}

\usepackage{algorithm}
\usepackage[noend]{algpseudocode}
\usepackage{amsmath}
\usepackage{multirow}
\usepackage{booktabs}
\usepackage[toc,page]{appendix}

\title{Imitating Unknown Policies via Exploration}

\addauthor{Nathan Schneider Gavenski\footnotemark[1]}{nathan.gavenski@edu.pucrs.br}{1}
\addauthor{Juarez Monteiro\footnotemark[1]}{juarez.santos@acad.pucrs.br}{1}
\addauthor{Roger Granada}{roger.granada@acad.pucrs.br}{1}
\addauthor{Felipe Meneguzzi}{felipe.meneguzzi@pucrs.br}{1}
\addauthor{Rodrigo C. Barros}{rodrigo.barros@pucrs.br}{1}

\addinstitution{
    Machine Learning Theory and Applications Research Group (MALTA), \\
    School of Technology,\\
    Pontif{\'i}cia Universidade \\
    Cat{\'o}lica do Rio Grande do Sul,\\
    Porto Alegre, RS, Brazil
}

\runninghead{Gavenski et al.}{Imitating Unknown Policies via Exploration}

\def\eg{\emph{e.g}\bmvaOneDot}

\def\etal{\emph{et al}\bmvaOneDot}

\def\ie{\emph{i.e}\bmvaOneDot}
\def\etc{\emph{etc}\bmvaOneDot}
\def\Tau{\mathcal{T}}

\begin{document}

\maketitle
\footnotetext[1]{These authors contributed equally to the work.}
\footnotetext[2]{The official implementation of IUPE is available at: https://github.com/NathanGavenski/IUPE}
\begin{abstract}

Behavioral cloning is an imitation learning technique that teaches an agent how to behave through expert demonstrations.
Recent approaches use self-supervision of fully-observable unlabeled snapshots of the states to decode state-pairs into actions.
However, the iterative learning scheme from these techniques are prone to getting stuck into bad local minima.
We address these limitations incorporating a two-phase model\footnotemark[2] into the original framework, which learns from unlabeled observations via exploration, substantially improving traditional behavioral cloning by exploiting (i) a sampling mechanism to prevent bad local minima, (ii) a sampling mechanism to improve exploration, and (iii) self-attention modules to capture global features. 
The resulting technique outperforms the previous state-of-the-art in four different environments by a large margin. 
\end{abstract}

\section{Introduction}
\label{sec:introduction}
                        
Humans often learn new abilities by observing other humans performing certain activities: we can mirror the behavior of others by observing sequences of events even though we may not have direct access to the underlying actions and intentions that yielded them~\cite{RizzolattiSinigaglia2010}. 
For example, we can learn different tasks (say cooking or woodworking) just by watching videos. 
Human capabilities go beyond merely imitating; we can transfer the knowledge from a demonstration to a given task, despite differences in the environment, body, or objects that constitute the demonstration.

Imitation learning~\cite{RazaEtAl2012}, also referred to as~\textit{learning from demonstration} (\textit{LfD}), consists of an artificial agent mimicking the behavior of an expert by learning from demonstrations~\cite{Schaal1996,ArgallEtAl2009}.
In LfD, the agent is trained to acquire skills or behaviors by observing a teacher solving a  problem, \ie, the agent learns a mapping between observations and actions~\cite{HusseinEtAl2017,TorabiEtAl2018}. 
While LfD is motivated by how humans learn from observation, most existing work assumes the agent has access to the actions (\ie, action labels) being performed, which humans do not. 
This assumption differs from how humans learn since they efficiently decode the observed information into the underlying actions~\cite{RizzolattiSinigaglia2010}.
Using labeled actions to learn a policy often requires the data to be recorded explicitly for imitation learning, limiting the usage of available unlabeled data.
Such an assumption is restrictive and unrealistic since usually we do not have direct access to the label of the action that is being performed by the expert.
Recent approaches for overcoming those issues perform \textit{imitation from observation} (IfO). 
In IfO, an artificial agent learns policies by using only the sequence of state observations, \ie, without the need to access the demonstrator's actions~\cite{LiuEtAl2018,TorabiEtAl2018,aytar2018playing}. 

In this paper, we address the problem of IfO by learning to imitate the behavior of an expert through the use of its state information without any other prior information of the observed actions. 
Our approach, \textit{Imitating Unknown Policies via Exploration} (IUPE), substantially improves over traditional behavior cloning by exploiting three novel ideas that we develop in this paper: (i)~the impact of a carefully-designed sampling mechanism that regulates the observations that are used to feed the inverse dynamics model, preventing the models from reaching bad local minima (Sec.~\ref{subsec:sampling}); (ii)~the impact of sampling from the softmax distribution of actions instead of reaching for the maximum \textit{a posteriori} estimation, as a means to also add stochasticity (\ie, improve \textit{exploration}) to the iterative imitation learning process and prevent local minima  (Sec.~\ref{subsec:exploration}); and~(iii)~the impact of self-attention  (Sec.~\ref{subsec:attention}) within both the inverse dynamics model and the policy model. 
By making use of \textit{sampling}, \textit{exploration}, and \textit{self-attention}, IUPE is capable of outperforming all state-of-the-art approaches based on behavior cloning, either over low-dimensional state spaces or over raw images, regarding both \textit{performance} and \textit{Average Episodic Reward} ($AER$).

\section{Related Work}
\label{sec:related_work}

Many approaches for imitating from observations have been recently proposed. 
A \emph{generative adversarial imitation learning} (GAIL)~\cite{HoErmon2016} approach learns to imitate policies from state-action demonstrations using adversarial training \cite{GoodfellowEtAl2014}. 

Pavse \etal~\cite{PavseEtAl2019} propose the reinforced inverse dynamics modeling (RIDM) method that combines reinforcement learning and imitation from observation to perform imitation using a single expert demonstration. 
Edwards \etal~\cite{EdwardsEtAl2019} describe a method called \emph{imitating latent policies from observation} (ILPO), where the agent first learns a latent policy offline that estimates the probability of a latent action given the current state. 
Then, in a limited number of steps in the environment, they perform remapping of the actions, associating the latent actions to the corresponding exact actions. 
Torabi \etal~\cite{TorabiEtAl2018} develop \emph{behavioral cloning from observation} (BCO) that uses a \emph{inverse dynamics model} to learn the meaning between the state's transition (action) and a \emph{policy model} to imitate the expert's behavior.
Its augmented version (ABCO)~\cite{monteiro2020augmented} improves on the sampling mechanism from BCO by partially sampling from the policy experiences.

\section{Imitating Unknown Policies via Exploration}
\label{sec:iupe}

Our problem assumes an agent acting in a Markov Decision Process (MDP)  represented by a five-tuple $M = \{S, A, T, r, \gamma\}$ \cite{SuttonBarto1998}, in which:  $S$ is the state-space, $A$ is the action space, $T$ is the transition model, $r$ is the immediate reward function, and $\gamma$ is the discount factor. 
Solving an MDP yields a stochastic policy $\pi(a|s)$ with a probability distribution over actions for an agent in state $s$ that needs to take a given action $a$.
\emph{Imitation from observation}~\cite{TorabiEtAl2018}, aims to learn the inverse dynamics $\mathcal{M}_{a}^{s_t,s_{t+1}} = P(a|s_t,s_{t+1})$ of the agent, \ie, the probability distribution of each action $a$ when the agent transitioned from state $s_t$ to $s_{t+1}$. 
In this problem, the reward function is not explicitly defined, and the actions performed by the expert are unknown, so we want to find an imitation policy from a set of state-only demonstrations of the expert $D = \{\zeta_1, \zeta_2, \ldots, \zeta_N\}$, where $\zeta$ is a state-only trajectory $\{s_0, s_1, \ldots, s_N\}$.

\emph{Imitating Unknown Policies via Exploration} (IUPE) follows the Behavioral Cloning from Observations (BCO) framework~\cite{TorabiEtAl2018}, further augmenting it with two strategies for avoiding local minima, \textit{sampling} and \textit{exploration}, and with self-attention modules for improving the learning of global features and, hence, generalization.
Algorithm~\ref{alg:iupe} summarizes the IUPE training process, where $\mathcal{I}^{pre} = \{(s_{t}^a, a_t, s_{t+1}^a)\}$ contains pre-demonstrations that are generated using a random policy $\pi$, $\Tau^{e} = \{(s^e_t, s^e_{t+1})\}$ contains pairs of states from expert demonstrations $D$, $\mathcal{I}   ^{pos} = \{(s_{t}^e, \hat{a}_t, s_{t+1}^e)\}$ contains post-demonstrations with states from the expert and their predicted actions, $v_e$ is a value set to 1 if the agent achieves the environment goal or zero otherwise, and $E$ is the set of runs in an environment. 

\begin{algorithm}[!hbt]
    \caption{High-level IUPE framework.}
    \begin{algorithmic}[1]
        \State Initialize model $\mathcal{M}_\theta$ as a random approximator
        \State Initialize policy $\pi_\phi$ with random weights
        \State Generate $\mathcal{I}^{pre}$ using policy $\pi_\phi$
        \State Generate state transitions $\Tau^{e}$ from demonstrations $D$
        \State Set $\mathcal{I}^s = \mathcal{I}^{pre}$
        \State Let $\alpha$ be the number of improvement cycles
        \For { $i \gets 0$ to $\alpha$ }
            \State Improve $\mathcal{M}_{\theta}$ by trainIDM($\mathcal{I}^s$)
            \State Use $\mathcal{M}_{\theta}$ with $\Tau^{e}$ to predict actions $\hat{A}$
            \State Improve $\pi_\phi$ by behavioralCloning($\Tau^{e}$, $\hat{A}$)
            
            \For { $e \gets 1$ to $\left|E\right|$ }
                \State Use $\pi_\phi$ to solve environment $e$ 
                \State Append samples $\mathcal{I}^{pos} \gets (s_t, \hat{a}_t, s_{t+1})$
                \If { $\pi_\phi$ at goal $g$}
                    \State Append $v_e \gets 1$
                \Else 
                    \State Append $v_e \gets 0$
                \EndIf
            \EndFor
            \State Set $\mathcal{I}^s =$ sampling($\mathcal{I}^{pre}$, $\mathcal{I}^{pos}$, $P(g|E)$, $v_e$)
        \EndFor
    \end{algorithmic}
    \label{alg:iupe}
\end{algorithm}

\subsection{Behavioral Cloning from Observation}
\label{subsec:bco_framework}

Behavioral Cloning from Observation~\cite{TorabiEtAl2018} is a framework that comprises a module that learns the inverse dynamics of the environment (Inverse Dynamics Model, IDM), and a module that learns an imitation policy.
The IDM is responsible for learning the actions that make the agent transition from state $s_t$ to $s_{t+1}$. 
Before training the IDM, the agent interacts with the environment via a random policy $\pi$ and generates pairs of states and the corresponding action, saving them as pre-demonstrations $\mathcal{I}^{pre} = \{(s_{t}^a, a_t, s_{t+1}^a)\}$.
The IDM uses the pre-demonstrations to learn the inverse dynamics $\mathcal{M}_\theta$ of the agent by finding parameters $\theta^*$~that best describes the state transitions. 

The policy model (PM), in turn, is responsible for cloning the expert's behavior. 
Based on pairs of states from expert demonstrations $\Tau^{e} = \{(s^e_t, s^e_{t+1})\}$, the IDM is employed to compute the predicted distribution over actions $\mathcal{M}_\theta (s_t^e, s_{t+1}^e)$ and predict the action $\hat{a}$ that corresponds to the movement taken by the expert to change from state $s_t$ to $s_{t+1}$. 
With the states from the expert and the predicted actions (self-supervision) for each state, the method learns the imitation policy by finding parameters $\phi^*$ for which $\pi_\phi$ best matches the provided tuples $\{(s_t^e, \hat{a})\}$. 
After learning the imitation policy, one creates a new dataset of post-demonstrations $\mathcal{I}^{pos} = \{(s_{t}^e, \hat{a}_t, s_{t+1}^e)\}$ with states from the expert and predicted actions. 

Both IDM and PM can learn in an iterative process known as BCO($\alpha$), where $\alpha$ represents a parameter to control the number of post-demonstration interactions. 
In BCO($\alpha$), the IDM uses post-demonstrations generated by the PM to update its model, and then the PM is updated using the new outcomes from the updated IDM. 
Since the original BCO($\alpha$) uses only the set of post-demonstrations to re-train the IDM, the predictive performance of the IDM tends to degrade when post-demonstrations contain misleading actions for specific pairs of states, which is often the case in the initial iterations. 
We address this problem by weighing how much the IDM should learn from pre-demonstrations ($\mathcal{I}^{pre}$) and how much from post-demonstrations ($\mathcal{I}^{pos}$), as detailed next.

\subsection{Sampling}
\label{subsec:sampling}

For obtaining the samples from post-demonstrations, we first select the distribution of actions given an environment and the current policy $P(A|E;\mathcal{I}^{pos})$.
Our sampling strategy considers only successful runs from $\mathcal{I}^{pos}$, \ie, only state-action sequences in which the agent was able to achieve the environment goal.
We represent it as $v_e$ in Eq.~\ref{eq:expert_distribution}, where $v_e$ is set to 1 if the agent achieves the environment goal or zero otherwise, and $E$ is the set of runs in an environment:

\begin{equation}
    \begin{split}
    \label{eq:expert_distribution}
    P(A|E;\mathcal{I}^{pos}) = \frac{\sum_{e=1}^{|E|} v_e \times  P(A|e)}{|E|}
    \end{split}
\end{equation}

The only assumption we make of the available demonstration data is that most image sequences are those of the demonstrator achieving a goal, even if such demonstrations have no label indicating whether the demonstration achieves the goal or not.
The intuition for using the post-demonstrations from successful runs alone is that if a policy is not able to achieve the environment goal, then the post-demonstration alone is not enough to close the gap between what the model previously learned with $\mathcal{I}^{pre}$ and what the expert performs within the environment. 
Using only successful runs also gives us a more accurate distribution of the expert's actions since we are only using those observations that achieved the objective instead of the random performance that maps the dynamics of the agent within a balanced dataset of actions. 
By not adding unsuccessful runs to the training dataset, we solve the problem in which BCO($\alpha$) degrades the performance in both models. 
With the distribution of actions for successful runs, we select all samples $\mathcal{I}_{spl}^{pos}$ from those runs, assuming each action follows the distribution from Eq.~\ref{eq:expert_distribution}.

The next step is to sample from the pre-demonstrations $\mathcal{I}_{spl}^{pre}$ with the inverse probability of the post-demonstrations, \ie, the loss probability distribution given by $1- P(A|E;\mathcal{I}^{pos})$. In a nutshell, the samples (observations) that comprise the novel post-demonstration dataset are sampled proportionally to $P(A|E;\mathcal{I}^{pos})$ for winning executions, and the dataset is filled with the pre-demonstrations proportional to the number of losses.
The final set $\mathcal{I}^{s}$ contains the concatenation of both $\mathcal{I}_{spl}^{pre}$ and $\mathcal{I}_{spl}^{pos}$ samples.

Complementing the training dataset with random demonstrations has two main advantages.
First, it helps the model to avoid overfitting the policy demonstrations.
Second, in the early iterations, when the policy generates very few successful runs, and its action distribution is entirely dissimilar to the expert, the training data will guarantee improved exploration by the IDM. With the win-loss probability, we force the training data to be closer to the expert demonstration than to the random data, which boosts the model capability of imitating the expert. 
The full sampling pseudocode can be found in the supplementary material. 

Note that our method is only goal-aware and does not use any reward information for learning or optimizing the models. 
We do not want to use rewards since not all environments provide intuitive reward functions for solving the problem. 
On the other hand, most agents have a clear goal that is relatively easy to visually identified. Examples include: (i)~the \emph{mountaincar} reaching the flag pole; (ii)~arriving at the final square at a given \emph{maze}; (iii)~\emph{acrobot} reaching the horizontal line; and (iv)~the \emph{cartpole} surviving up to 195 steps. 

\subsection{Exploration}
\label{subsec:exploration}

The original behavioral cloning framework uses the \textit{maximum a posteriori} (MAP) estimation, \ie, it predicts the action with the greatest probability given by the model for a pair of states, both in its original version and in its $\alpha$-iterations version.
By using MAP predictions, we identified several cases in which the model is still relatively unsure about the correct action, especially in earlier iterations, leading to undesired local minima.

We borrow a simple solution from the area of language modeling to avoid such a problem, which is to sample the actions from the softmax distribution of both models (IDM during the expert labeling and PM during the execution of the environment). 
By not using the MAP estimation, we create a stochastic policy capable of further exploration in early iterations considering the model uncertainty. 
We show that creating those samples allows the IDM to converge in fewer iterations, since the sampling method guarantees a more sparse dataset consisting of $\mathcal{I}^{pre}$ and $\mathcal{I}^{pos}$, and the stochastic policy guarantees more exploration of the search space for properly achieving the environment goal.

Furthermore, in dynamic environments, the stochastic policy contributes to reaching the goal when deterministic behavior would not.
This difference is vital for avoiding local minima during iterations.
If the model were not capable of sampling a sub-optimal action during its training phase, the agent actions would resume for the most common action in the expert samples.
By sampling the most frequent action, the policy is susceptible to looping between states, \eg, choosing left and right interchangeably.

\subsection{Self-Attention}
\label{subsec:attention}

Self-Attention (SA) \cite{VaswaniEtAl2017,ZhangEtAl2019} is a module that learns long-range dependencies within the internal representation of a neural network by computing non-local responses as a weighted sum of the features at all positions. 
It allows the network to focus on specific features that are relevant to the task at each step. 
Self-Attention in IUPE minimizes the impact of the constant changes in the post-demonstrations at each iteration. 
The smaller impact occurs due to both models weighting all features, thus disregarding the potential local noise that the agent might create and improving the final classification. 
Given that in the early iterations of IUPE the PM is rarely capable of achieving the goal, the IDM has time to learn with the pre-demonstration how an action looks with no domain expertise. 
As the PM gets better, the IDM is capable of fine-tuning its classification power to predict the movement from the expert accurately. 
During backpropagation, both models have smoother weight updates as a consequence of the weighting of all features enabled by the SA module.

The SA module can also be applied to image representations~\cite{ZhangEtAl2019}. In such high dimensional space, it learns to correlate regions that are not present in the same filter of a convolutional layer. 
When applied to images, the self-attention module is capable of identifying which part of the states is essential for predicting the correct action.


\section{Experimental Results}
\label{sec:experiments}

In this section, we describe the details of the IUPE algorithm, the networks we use in each environment, the metrics for evaluating the results, and our findings in comparison with the state-of-the-art.
Our experiments are based on 4 different environments from OpenAI Gym \cite{brockman2016openai}, where we separate vector-based environments -- \emph{Acrobot-v1}, \emph{Cart-Pole-v1}, \emph{MountainCar-v0} -- from image-based environments -- \emph{Gym-Maze} $3\times3$, $5\times5$, and $10\times10$.
All experiments, including the baselines, run for 100 epochs sharing the same expert data.

\subsection{Implementation and Metrics}
We create two different networks to address each type of environment: a network for low-dimensional \emph{vector-based environments}, and a network for high-dimensional \emph{image-based environments}.
All networks are implemented on \emph{PyTorch} and optimize the cross-entropy loss function via Adam~\cite{kingma2014adam}. 
We add self-attention~\cite{VaswaniEtAl2017,ZhangEtAl2019} modules in both IDM and PM.

We evaluate IUPE and the related work in terms of both \textit{Average Episodic Reward} ($AER$) and \textit{Performance} ($P$) metrics. 
$AER$, displayed in Equation~\ref{eq:aer}, is the average reward of $E$ runs for each environment considering the total amount of footsteps ($F_i$).
Considering that $AER$ is dependent on an environment reward function, its value differs from task to task. 
$AER$ measures how well the expert performs the task, indicating how difficult it is for the agent to imitate the expert's behavior. 
We calculate $AER$ by averaging a 100 different Mazes for the Gym-maze environment, and 100 consecutive runs for the other environments. 

\begin{equation} \label{eq:aer}
    AER = \frac{\sum_{i=1}^{E}\sum_{j=1}^{F_i}\pi_\phi(e_{ij})}{E}
\end{equation}

\emph{Performance}, exemplified in Equation~\ref{eq:performance}, is the average reward for each run scaled from 0 to 1, where zero is the random policy ($\pi_\xi$) reward, and one is the expert ($\pi_\varepsilon$).
A model can achieve scores lower than zero if it performs worst than random and higher than one if it performs better than the expert. 
We do not use accuracy, since achieving high accuracy in $\mathcal{I}^s$ does not guarantee good results on solving the tasks. 
The accuracy of the PM highly correlates with the IDM, which does not carry domain-specific information. 
When achieving 100\% of accuracy with the PM, the model may have entirely learned the IDM behavior and its errors, propagating them during inference and yielding low $AER$ and $P$ scores.

\begin{equation} \label{eq:performance}
    P = \frac{\sum_{i=1}^{E}\frac{\pi_\phi(e_i)-\pi_\xi(e_i)}{\pi_\varepsilon(e_i) - \pi_\xi(e_i)}}{E}
\end{equation}

\subsection{Results}
\label{sec:results}

\begin{table*}[tb!]
  \centering
  \scriptsize
  \begin{tabular*}{\textwidth}{@{\extracolsep{\fill}} c c  r r r r r r}
    \toprule
        \multicolumn{1}{c}{Models} & \multicolumn{1}{c }{Metrics} & CartPole & Acrobot & MountainCar & Maze $3\times3$ & Maze $5\times5$ & Maze $10\times10$\\
    
    \midrule
    
    \multirow{2}{*}{Expert} 
     & $P$                       & $1.000$   & $1.000$ & $1.000$ & $1.000$ & $1.000$ & $1.000$\\
     & $AER$                     & $442.628$ & $-110.109$ & $-147.265$ & $0.963$ & $0.970$ & $0.981$\\
    \midrule
    
    \multirow{2}{*}{Random} 
     & $P$                       & $0.000$  & $0.000$    & $0.000$    & $0.000$ & $0.000$ & $0.000$\\
     & $AER$                     & $18.700$ & $-482.600$ & $-200.000$ & $0.557$ & $0.166$ & $-0.415$\\
    
    \midrule

    \multirow{2}{*}{BC} 
     & $P$                       & $1.135$   & $1.071$   &  $1.560$   & $-1.207$ & $-0.921$ & $-0.470$\\
     & $AER$                     & $500.000$ & $-83.590$ & $-117.720$ & $0.180$  & $-0.507$ & $-1.000$\\
    
\midrule
\midrule
    
    \multirow{2}{*}{BCO}
     & $P$                       & $\mathbf{1.135}$   & $0.980$    & $0.948$    & $0.883$ & $-0.112$ & $-0.416$\\
     & $AER$                     & $\mathbf{500.000}$ & $-117.600$ & $-150.000$ & $\mathbf{0.927}$ & $0.104$  & $-0.941$\\
    \midrule
    
    \multirow{2}{*}{ILPO}  
     & $P$                       & $\mathbf{1.135}$   & $1.067$   &	$0.626$	& $-1.711$ & $-0.398$ &	$0.257$\\
     & $AER$                     & $\mathbf{500.000}$ & $-85.300$ & $-167.000$ & $-0.026$ & $-0.059$ & $-0.020$\\
    \midrule
    
    \multirow{2}{*}{IUPE} 
     & $P$                       & $\mathbf{1.135}$   & $\mathbf{1.086}$   & $\mathbf{1.314}$    & $\mathbf{1.361}$ & $\mathbf{1.000}$ & $\mathbf{1.000}$\\
     & $AER$                     & $\mathbf{500.000}$ & $\mathbf{-78.100}$ & $\mathbf{-130.700}$ & $\mathbf{0.927}$ & $\mathbf{0.971}$ & $\mathbf{0.981}$\\
    \bottomrule
\end{tabular*}
\caption{\emph{Performance} and \emph{Average Episode Reward} for our approach and related work.}
\label{tab:results}
\end{table*}

We evaluate our approach by comparing it with the state-of-the-art approaches in behavioral cloning: BCO~\cite{TorabiEtAl2018} and ILPO~\cite{EdwardsEtAl2019}). 
Table~\ref{tab:results} shows the results achieved by each method. 
For comparison purposes, we also show the results (in terms of both $AER$ and $P$) for the expert, for a random policy and for behavioral clone (BC), which is a supervised approach. 
All models are trained using the same initial set of random pre-demonstrations $\mathcal{I}^{pre}$.

{\bf Overall Results:} As shown in Table~\ref{tab:results}, IUPE outperforms the state-of-the-art approaches for all environments but \emph{CartPole}, where it provides the same results as the baselines.
Results in CartPole are expected since it contains only four dimensions to represent the state, and the actions are easily separated.
In Acrobot, IUPE and ILPO achieve similar \emph{Performance} ($P\approx1.00$), showing that both approaches have similar imitation abilities. 
However, IUPE achieves $AER=-78.10$ while ILPO obtains $-85.30$, which means that IUPE can solve the environment with an advantage of $\approx10$ frames in advance than ILPO.
For MountainCar, IUPE achieves $P=1.314$, which is the best result when compared with the baselines in this environment, a difference of $\approx0.37$ from the best second result (BCO).
$AER$ values for MountainCar are also much better for IUPE: $-130.70$, which is $19.3$ better than BCO.
For the Maze environments, IUPE reaches top $P$ values for all types of Mazes that were tested.
In terms of $AER$, IUPE was only reached by BCO in Maze $3\times3$, where both models achieve $AER=0.927$.
IUPE is virtually unaffected in the same environment as the complexity of the Maze increases, whereas the other approaches fail to learn the dynamics. 
IUPE can learn due to the stochasticity that gives the model the ability to better explore the environment, avoiding local minima and reaching the goal at the end. 
Compared with the \emph{Expert}, IUPE can achieve similar or better metric values for most environments.

{\bf IUPE \textit{vs.} BCO:} Since IUPE is an augmentation of BCO, we can observe in the results that the augmentation affects the imitation process positively. 
In all environments, IUPE achieves similar or better results when compared with BCO. 
IUPE and BCO obtained the same $AER$ score in Maze $3\times3$ due to both methods resolving the same number of environments. 
However, IUPE was able to achieve higher rewards than BCO in the same environment. 
Unlike BCO, IUPE keeps learning as the environment's number of states increases.

{\bf IUPE \textit{vs.} ILPO:} ILPO provides similar results as IUPE for the CartPole and Acrobot environments. However, IUPE achieves better $P$ and $AER$ scores for MountainCar and all the Mazes. 
In MountainCar, IUPE obtained $P=1.314$, which means that it was able to learn how to imitate the expert and even surpass it. 
In terms of $AER$, IUPE achieves $-130.7$ versus $-167$ from ILPO, meaning that for 100 episodes, our approach achieves better rewards in MountainCar. 
In the Maze environments, ILPO achieves $AER$ scores below zero, which means that it could not solve most of the mazes. 
We believe that ILPO achieves low scores because it does not consider the full image of the scenario since it uses crop mechanisms and internal manipulations with the state images. 
Such mechanisms could lead to training data without essential aspects in the images (\eg~the agent, the goal state, \etc).


\section{Discussion}
\label{sec:discussion}

We now briefly discuss the three key strategies we explore in the context of behavioral cloning from observations: \emph{self-attention}, \emph{sampling}, and \emph{exploration} over \emph{maximization}. 
We run ablation studies to analyze the impact of each feature separately, as well as combined with the other features. 
Table~\ref{tab:ablation} shows the results of this analysis for the $10\times 10$ Maze environment, where BCO represents the approach without considering any improvement.

\begin{table}[!t]
    \centering
    \scriptsize
    \begin{tabular*}{\columnwidth}{@{\extracolsep{\fill}}lrr} 
        \toprule
        Model & $P$ & $AER$ \\ 
        \midrule
        BCO                                       & $-0.416$         &  $-0.941$ \\
        Attention                                 & $-0.415$         &  $-0.940$ \\
        Sampling                                  & $0.534$          &  $0.348$ \\
        Exploration                               & $0.734$          &  $0.605$ \\
        Attention + Sampling                      & $0.367$          &  $0.088$ \\ 
        Attention + Exploration                   & $-0.407$         &  $-0.921$ \\
        Sampling + Exploration                    & $0.943$          &  $0.901$ \\
        Attention + Sampling + Exploration (IUPE) & $\mathbf{1.000}$ &  $\mathbf{0.981}$ \\ 
        \bottomrule
    \end{tabular*}
    \caption{Ablation study considering IUPE's 3 main components in the maze environment.}
    \label{tab:ablation}
\end{table}

\begin{figure}[!b]
    \centering
    \includegraphics[width=0.95\textwidth]{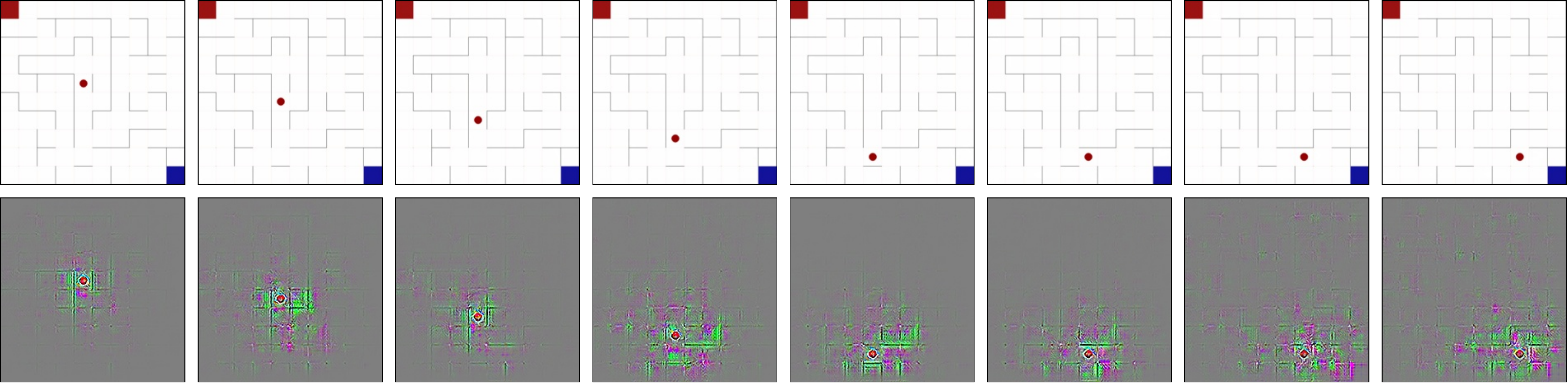}
    \caption{Heatmap visualization of the gradient filters activating for the maze environment. The first row shows the input image, while the second row shows the gradient activation in the first self-attention module.}
    \label{fig:gradcam}
\end{figure}

\begin{figure}[t!]
    \centering
    \subfigure[]{\label{fig:distribution}\includegraphics[width=0.4\textwidth]{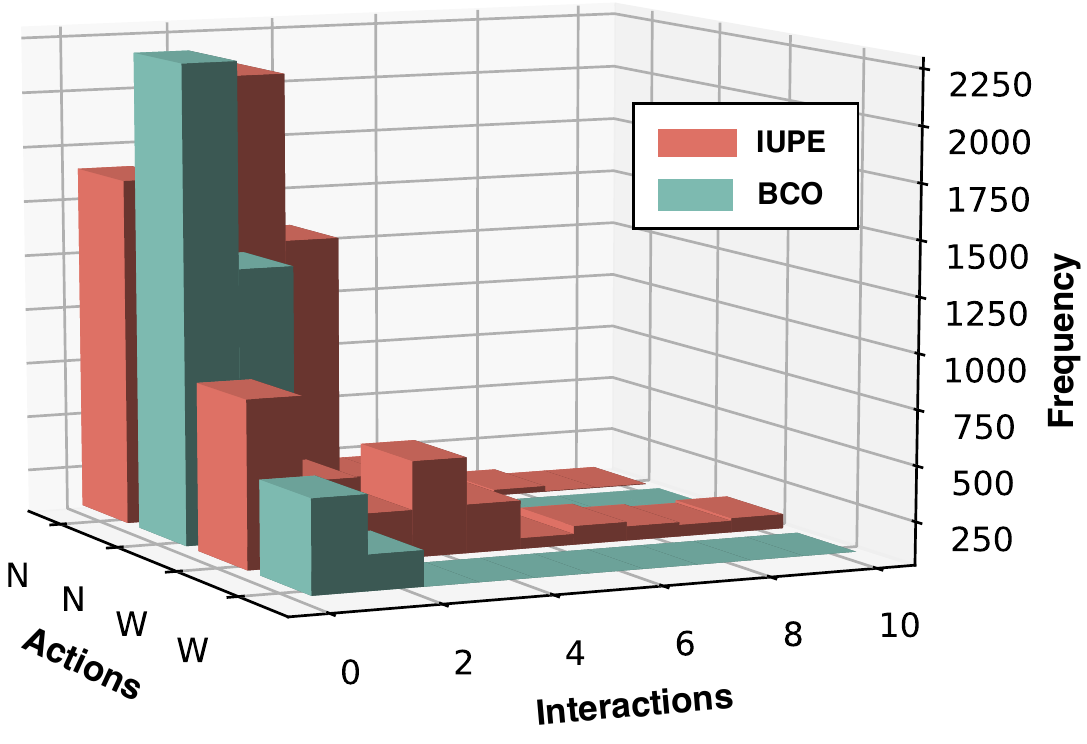}}
    \hspace{0.05\textwidth}
    \subfigure[]{\label{fig:exploration_decay}\includegraphics[width=0.4\textwidth]{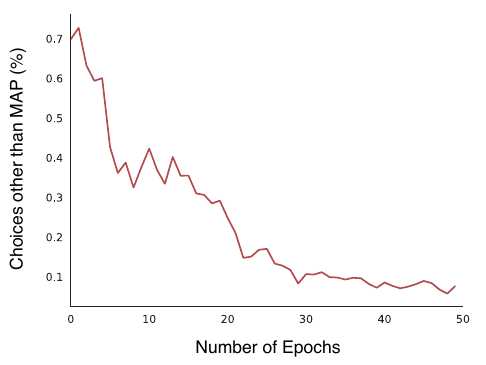}}
    \caption{(a) IDM predictions for the expert examples through time. 
    (b) Percentage of choices in which the MAP estimation is not selected by the self-decaying exploration rate.}
    \label{fig:imagem}
\end{figure}

{\bf Self-Attention:} By using the SA modules alone, IUPE achieves similar results to BCO in terms of $P$ and $AER$, though it converges earlier than BCO in the training phase. 
We observe that the SA modules have a larger impact when shifting from $\mathcal{I}^s$ to $\mathcal{T}^e$ due to the weighting of all features that the attention provides, maintaining high accuracy in between epochs.
This behavior can be explained in Fig.~\ref{fig:gradcam}, where we used the Grad-CAM~\cite{selvaraju2017grad} technique to visualize the self-attention gradient activations given an image in a trained policy.
By observing the gradient activations, we conclude that the self-attention modules help the model to focus on the agent, while still being able to pay attention to walls nearby.
Furthermore, activation areas are wider when the agent is walking through open passages than between corridors.
We present a video representation of the agent's gradients activation in the supplementary material.
This behavior resembles human vision and is exemplified in the first and third frames, where the agent is between two walls, while in the other frames the agent has a broader view.
However, by not having the sampling method and using the original BCO reconstruction for the IDM dataset, we face the problem of actions vanishing from the model prediction distribution, as illustrated in Fig.~\ref{fig:distribution}.
This problem may occur when the PM does not execute all actions, or when it acts very differently from the expert, causing the feature space to be far from the expert actions.

Our method achieves a lower variance in scores during the validation phase compared to BCO, indicating that the model learned how to better identify the correct action from the tuple of states. 
After evaluating the impact of self-attention (Table~\ref{tab:ablation}) and the vanishing action issue, we experimented with the sampling mechanism that regulates the observations used to feed the IDM, preventing it from reaching bad local minima.

{\bf Sampling:} By inspecting Table~\ref{tab:ablation}, we can see that using the sampling method alone is enough to solve the problem of the vanishing actions and achieving a significant improvement from BCO and self-attention by itself.
Since the sampling mechanism keeps $\mathcal{I}^{pre}$ either integrally or partially throughout the iterations, it allows the IDM to keep receiving as input some random actions, which results in more balanced predictions.

As the training progresses and $\mathcal{I}^s$ progressively becomes more of $\mathcal{I}^{pos}$ than $\mathcal{I}^{pre}$, we can see a shift on the distribution of actions by the IDM that helps the PM to focus on actions that occur more frequently in the expert. 
We understand that this unbalancing of $\mathcal{I}^s$ helps the IDM to learn that the probability of all actions is not equal, and thus it should give more importance to some actions than others. 

When paired with the SA module, the sampling mechanism makes the PM achieve worse results when compared to the sampling method alone, though it still outperforms BCO.
We investigated the reason for that performance decay, and found out that the \textit{Attention+Sampling} model provided higher per-action accuracy during validation.
Though it had more certainty in which actions to choose, the agent was also much more likely to get stuck in between states, predicting an action that would send the agent to a previous state and creating a loop where the agent would go back and forth.

We thus noticed that the new sampling method alone can improve the learning experience in between iterations, but that when coupled with SA, it ended up decaying the generalization power of the model. Since the problem is in the excessive certainty regarding a few actions, we further experimented with the stochastic mechanism we called \textit{exploration}.

{\bf Exploration over Maximization:} Stochastic learning algorithms usually decay their exploration rate throughout iterations, since they assume that the agent progressively learns the optimal solution through time~\cite{mnih2015human}. 
If the decaying rate is too low, the agent might stick with the first solution it finds, and if it is too high it might spend too much time exploring sub-optimal states. 
By using the softmax distribution of actions, we create an exploration mechanism that naturally decays progressively as the neural network learns to separate the feature space (and hence there is no need for a decay hyperparameter to be tuned). 
The self-decaying exploration rate can be seen in Fig.~\ref{fig:exploration_decay}.

To understand whether IUPE benefits from this exploration rate, we can check Table~\ref{tab:ablation} and see the impact of exploration over the baseline (BCO), and also together with the remaining features (attention and sampling). 
By using this mechanism alone, we solve the problem of looping in between actions, and even though the PM may reach the goal through non-ideal paths, it allows the generation of samples that are closer to the expert than before. 

The exploration mechanism, when paired with the sampling method, achieves results similar to IUPE (full method with the 3 strategies).
This outcome is due to the increased stochasticity that helps in breaking action loops. While using exploration with SA results in a model with higher per-action accuracy, we perceive a decrease in both $P$ and $AER$. Our assumption is that this happens due to the action vanishing problem caused by the lack of the sampling method.
When all mechanisms are combined, we can see that the SA modules do not impact the model negatively anymore, but the opposite -- they slightly increase both $P$ and $AER$. 
With the exploration mechanism supporting the model and preventing it from getting stuck in between states, IUPE can fine-tune itself earlier due to the non-optimal actions being closer to the expert. 
By adding the sampling mechanism, which is responsible to balance $\mathcal{I}^s$, IUPE achieves the best results of this analysis.


\section{Conclusions and Future Work}
\label{sec:conclusions}
We developed a novel approach for imitation learning from observations with no prior information about the expert's actions. 
The pipeline of the architecture includes training an Inverse Dynamics Model and a Policy Model using a sampling mechanism that avoids local minima and using self-attention modules for capturing global features. 
Experiments show that our approach can use low-dimensional or image data to learn how to imitate, outperforming the current state-of-the-art methods.

As future work, we want to adapt our method to explore environments with continuous action spaces, and evaluate it in more challenging domains such as continuous control tasks, Atari games, and robotics goal-based tasks, among others. 
Such domains have larger state-spaces than our experimental environments, which makes them much harder to imitate. 
We plan to work with temporal approaches and with adversarial training in order to search for a model that is capable of imitating and perhaps generalizing to similar environments.

\paragraph{Acknowledgements} This study was financed in part by the Coordena\c{c}\~ao de Aperfei\c{c}oamento de Pessoal de N\'ivel Superior (CAPES) - Finance Code 001, and Funda\c{c}\~ao de Amparo \`a Pesquisa do Estado do Rio Grande do Sul (FAPERGS) agreement (DOCFIX 04/2018) process number 18/2551-0000500-2. We gratefully acknowledge the support of NVIDIA Corporation with the donation of the graphics cards used for this research.

\bibliography{bmvc_final}

\clearpage
\begin{appendices}

\section{Training flow}

\begin{figure}[!hbt]
    \centering
    \includegraphics[width=0.95\textwidth]{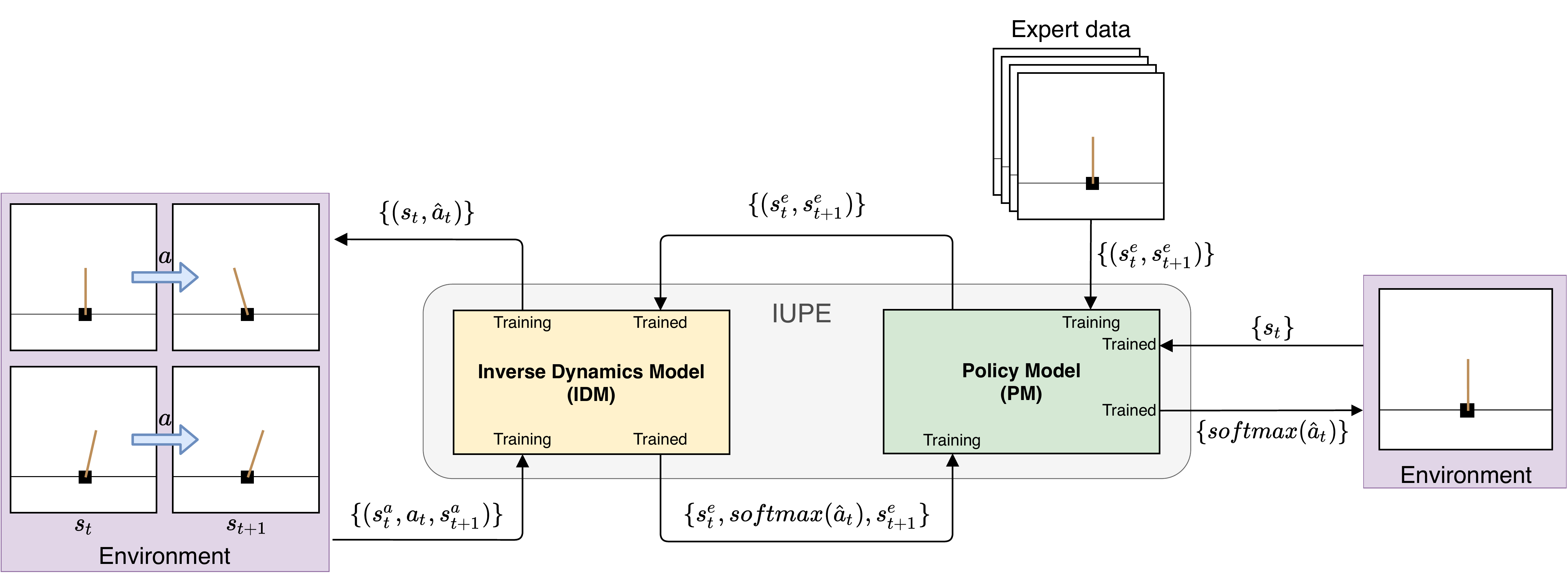}
    \caption{Pipeline representing the training flow for IUPE.}
    \label{fig:iupe_flow}
\end{figure}

In this section, we present the training flow for IUPE. As illustrated in Figure~\ref{fig:iupe_flow}, the process of training starts by collecting pairs of states ($s_t,s_{t+1}$) from a given environment, followed by a random action $a_t$ that was responsible for generating the transition between these two states.
Once the data is collected, we start to train the Inverse Dynamics Model, IDM,  as shown in the yellow box of Figure~\ref{fig:iupe_flow}. The IDM is responsible for predicting the action that caused the transition between two consecutive states.
Here, the IDM uses our proposed exploration strategy, which consists of choosing the action predicted by sampling from the $softmax$ distribution given the probabilities generated by the IDM.

The next step is to train the Policy Model, PM, as shown in the green box of Figure~\ref{fig:iupe_flow}.
In this part of the flow, we start by generating a label for the data provided by the expert.
For each pair of states ($s^e_t, s^e_{t+1}$) provided by the expert, the trained IDM will predict the action ($\hat{a}_t$) that caused such transition. 
After generating the new expert dataset, now containing a predicted action, we start to train the Policy Model.
The PM is trained by passing as input a given state ($s_t$), and the model needs to predict which action the expert will choose.
The error is calculated with the previous prediction that comes from the IDM.
In the end, the PM follows the same exploration strategy from the IDM. 
It samples from the $softmax$ distribution of the predictions from the PM to generate which action the agent needs to take.

\section{IUPE Models}

\begin{figure}[!htb]
    \centering
    \includegraphics[width=\textwidth]{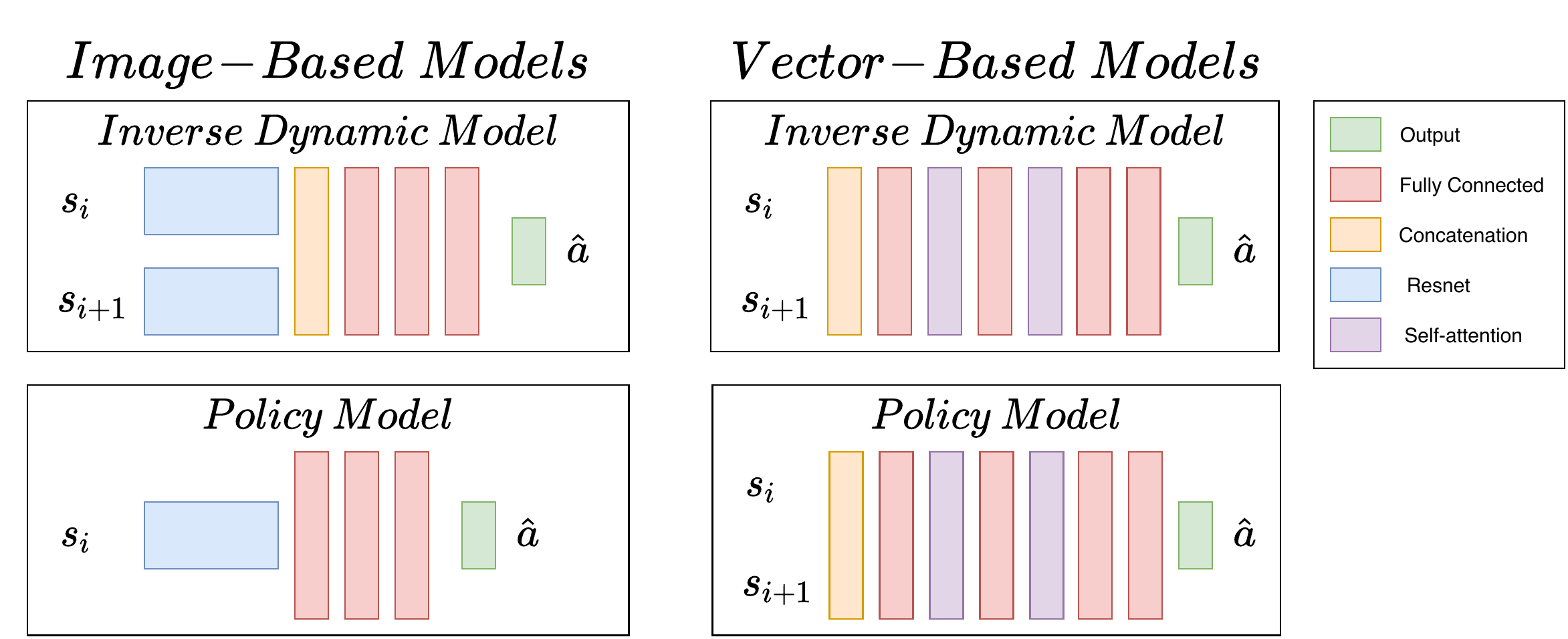}
    \caption{The Inverse Dynamics Model and the Policy Model representations for the image-based and vector-based environments. The concatenation layer for the image-based environments concatenates the visual embeddings from the ResNet encoder. For the vector-based environments, it concatenates the state representation.}
    \label{fig:iupe}
\end{figure}

In this section, we present the Inverse Dynamics Model and the Policy Model implementations that are used in this work. 
In Figure \ref{fig:iupe}, we divide the implementations in image-based and vector-based models given the environment they are on. Considering that the state representation varies in size, we do not present the input sizes. 
For the sake of simplicity, we show a simplified representation of the ResNet network.

\begin{figure}[!htb]
    \centering
    \includegraphics[width=0.75\textwidth]{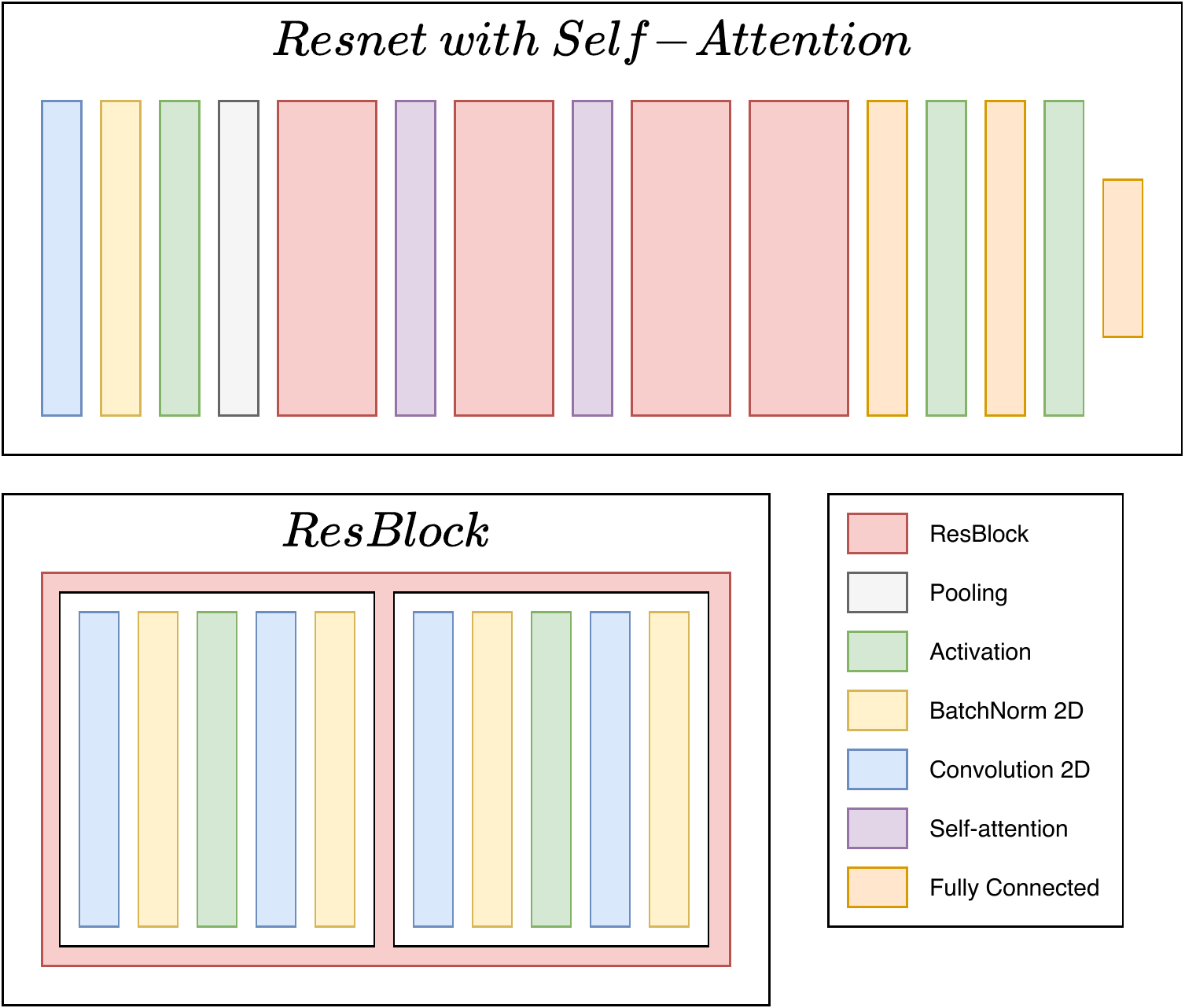}
    \caption{The ResNet model with self-attention modules. This implementation follows \textit{PyTorch}'s version with four ResBlocks. We add two self-attention modules to capture global features from the image state representations.}
    \label{fig:resnet}
\end{figure}

In Figure \ref{fig:resnet}, we present the self-attention modules that we add within the ResNet architecture. 
The overall architecture has been kept the same. We introduce both self-attention modules after the first and second ResBlocks from the encoder. The rest of the original implementation is untouched and follows the \textit{PyTorch}'s framework.

\end{appendices}

\end{document}